\documentclass{article}

\usepackage[preprint]{neurips_2026}

\usepackage[utf8]{inputenc}
\usepackage[T1]{fontenc}
\usepackage{hyperref}
\hypersetup{colorlinks=true,linkcolor=black,citecolor=blue,urlcolor=blue,pdftitle={Budgeted Quotient-Residual Guidance for Frozen Pocket-Conditioned Molecular Diffusion},pdfauthor={Xinyu Wang, Jinbo Bi, Minghu Song},pdfsubject={Machine Learning; Molecular Diffusion},pdfkeywords={quotient-residual guidance, molecular diffusion, frozen samplers}}
\usepackage{url}
\usepackage{booktabs}
\usepackage{multirow}
\usepackage{tabularx}
\usepackage{array}
\usepackage{amsfonts}
\usepackage{amsmath,amssymb,mathtools}
\usepackage{amsthm}
\usepackage{nicefrac}
\usepackage{microtype}
\usepackage{xcolor}
\usepackage{graphicx}
\usepackage{subcaption}
\usepackage{enumitem}
\usepackage{siunitx}
\usepackage{float}
\usepackage{placeins}
\usepackage[nameinlink,noabbrev]{cleveref}
\usepackage{xspace}
\graphicspath{{figures/}}

\newcommand{\budgetmethod}{\textsc{Budgeted Quotient-Residual Guidance}\xspace}
\newcommand{\qrg}{\textsc{QRG}\xspace}
\newcommand{\base}{\textsc{Base}\xspace}
\newcommand{\pcmain}{\textsc{PredNext-QRG}\xspace}
\newcommand{\cheapmain}{\textsc{Local-QRG}\xspace}
\newcommand{\teacher}{\textsc{Rollout-QRG Teacher}\xspace}
\newcommand{\sectiononly}{\textsc{Section-only}\xspace}

\newcommand{\sham}{\textsc{Sham-QRG}\xspace}

\newcommand{\TargetDiff}{\textsc{TargetDiff}\xspace}
\newcommand{\CBGBench}{\textsc{CBGBench}\xspace}
\newcommand{\GenBench}{\textsc{GenBench3D}\xspace}
\newcommand{\Vina}{AutoDock Vina\xspace}

\newcommand{\R}{\mathbb{R}}

\newcommand{\norm}[1]{\left\lVert #1 \right\rVert}
\newcommand{\inner}[2]{\left\langle #1,#2 \right\rangle}

\newtheorem{proposition}{Proposition}

\theoremstyle{remark}

\title{Budgeted Quotient-Residual Guidance for Frozen Pocket-Conditioned Molecular Diffusion}

\author{%
Xinyu Wang$^{1}$ \quad Jinbo Bi$^{1}$ \quad Minghu Song$^{2}$\\[4pt]
\normalfont\small $^{1}$Department of Computer Science and Engineering, University of Connecticut\\
\normalfont\small $^{2}$Institute of Health and Medicine, Hefei Comprehensive National Science Center\\
\normalfont\small $^{1}$Storrs, CT 06269, USA \quad $^{2}$Hefei 230601, China
}
\date{}

\begin{document}

\maketitle

\begin{abstract}
Pocket-conditioned molecular diffusion updates ambient atom coordinates, but many lead-optimization objectives are expressed on quotient features such as distances, contacts, and anchored substructures. We introduce \emph{budgeted quotient-residual guidance} (\qrg), an inference-time correction that makes these quotient objectives active without retraining the molecular generator. \qrg lifts quotient covectors to metric-horizontal ambient directions and delivers them through a trust budget set by the frozen sampler's own step norm: quotient geometry chooses the direction, while sampler motion bounds the scale. We derive the horizontal lift, closed-form sampler-budget update, KL/kinetic interpretation around a frozen reverse step, equivariance conditions, and a product-budget split for budget-capped section and residual controls. Controlled quotient tasks confirm that sampler-relative delivery activates signals that raw local quotient gradients leave dormant. On frozen \TargetDiff backbones, official seed-0 \CBGBench ligand-generation/editing sweeps show practical quality--runtime gains: \cheapmain improves validity from $0.815$ to $0.864$ on fragment growing, $0.664$ to $0.707$ on scaffold hopping, and $0.681$ to $0.712$ on linker design, while \pcmain improves fragment/scaffold and remains near-neutral on linker. Novelty remains $1.000$ and diversity is preserved in the matched multi-seed molecular slice, giving task-dependent improvements without sampler retraining or backbone modification. Overall, \qrg provides a lightweight route to quotient-aware inference for frozen molecular samplers with explicit runtime accounting.
\end{abstract}

\section{Introduction}

Pocket-conditioned molecular diffusion has two coordinate systems. The sampler evolves ambient atom coordinates, but many scientific objectives are defined only after quotienting away nuisance choices such as global rotation, translation, atom-index representatives, or irrelevant local coordinates. Fragment geometry, scaffold preservation, linker closure, and pocket-relative contacts are naturally described by distances, anchors, and substructure features rather than by one absolute coordinate representative. This ambient--quotient mismatch is present even when the backbone itself is equivariant: the model can move coordinates correctly under symmetry, while the guidance objective still lives on features that identify an equivalence class.

There are two broad ways to handle the mismatch. One can build the structure into training through equivariant architectures or symmetry-aware objectives \citep{thomas2018tensorfield,fuchs2020se3transformer,satorras2021egnn,hoogeboom2022edm,tong2025orbdiff}. This is principled, but it requires changing or retraining the generator. Alternatively, one can keep a pretrained sampler fixed and add plug-and-play guidance at inference time \citep{dhariwal2021diffusion,ho2022classifierfree,chung2023dps,bansal2023universal}. This second regime is attractive in structure-based drug design because strong pretrained backbones and benchmark code are already available, while retraining for every lead-optimization objective can be expensive or impractical \citep{guan2023targetdiff,lin2025cbgbench}.

Inference-time quotient guidance, however, has a failure mode that is easy to miss. A quotient energy removes nuisance directions, but the frozen sampler still takes finite ambient denoising steps. Direct local gradients of an invariant energy can be directionally meaningful yet numerically dormant: their scale is set by the quotient feature map rather than by the sampler motion. Rollout lookahead can activate such signals by repeatedly querying the denoiser, but then guidance becomes an expensive oracle rather than a practical inference rule. The missing ingredient is not another absolute gradient scale; it is a way to deliver a quotient direction at the scale already chosen by the sampler.

We propose \budgetmethod, a quotient-lifted trust-region formulation for frozen molecular diffusion samplers. The quotient differential first lifts a quotient covector into a metric-horizontal ambient direction, i.e., the ambient motion that changes quotient features most efficiently while ignoring vertical representative motion. The allowable correction length is then bounded by the frozen sampler's own update norm, which is also the local kinetic scale of the frozen reverse step. In one sentence, \emph{quotient geometry chooses direction; sampler motion bounds scale}. This converts a dormant invariant signal into an active, bounded correction while keeping the method compatible with pretrained backbones. The goal is not to replace training-time molecular generators or claim best-in-class de novo generation; it is to make quotient-aware inference useful when retraining the backbone is not the experiment one wants to run. Expensive rollout is demoted to a teacher or diagnostic, while deployed instantiations reuse either the sampler's predicted next state or a cheap local quotient surrogate.

\begin{figure}[t]
    \centering
    \includegraphics[width=\linewidth]{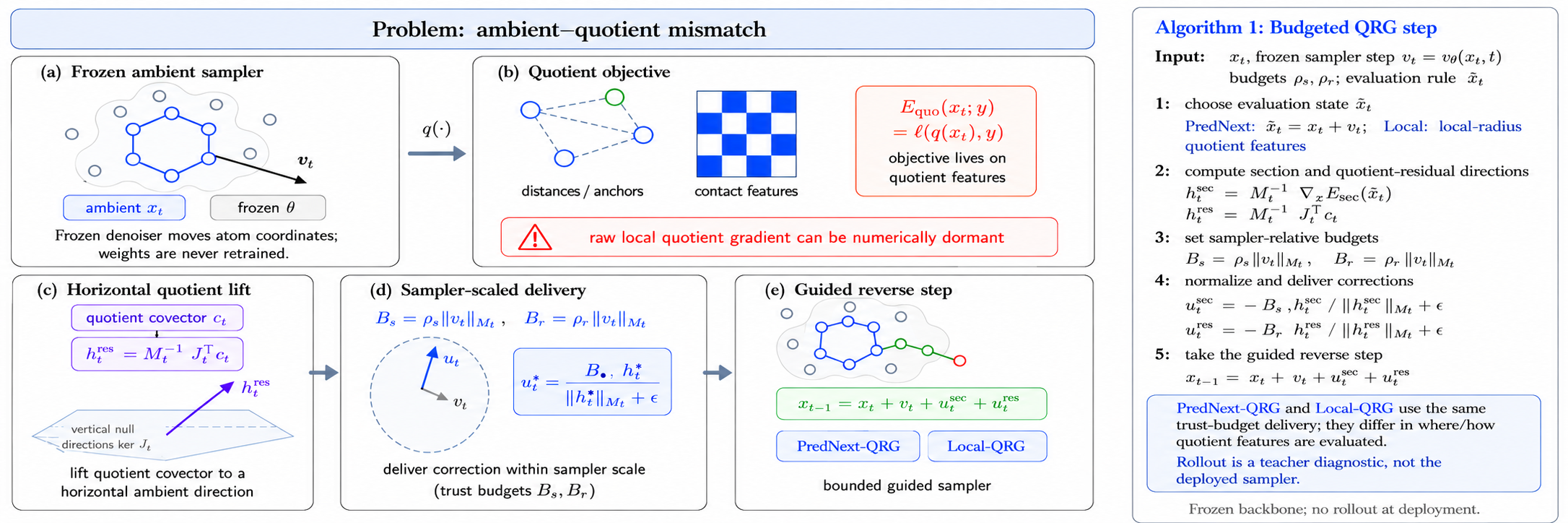}
    \caption{Budgeted quotient-residual guidance. A frozen molecular sampler moves ambient coordinates, while many scientific objectives live on quotient features such as distances, contacts, and fixed anchors. Raw quotient directions can be dormant at sampler scale. \qrg keeps the quotient-residual direction but delivers it through a sampler-relative trust budget. The right panel gives the deployed Budgeted QRG step: section and quotient-residual controls are budget-capped separately by the frozen sampler's own motion. Deployed variants use PredNext or local-radius quotient features; rollout residuals are teachers and diagnostics, not deployment requirements.}
    \label{fig:overview}
\end{figure}

Our empirical focus is the frozen-backbone setting. We integrate \qrg with \TargetDiff \citep{guan2023targetdiff} and evaluate on \CBGBench-style lead-optimization tasks \citep{lin2025cbgbench}. This choice makes the runtime question explicit: a useful inference-time method should improve generation quality without hiding a many-rollout cost. We therefore report validity, novelty, diversity, control budget, wall-clock runtime, paired target-level statistics, and corrected docking diagnostics. The main ligand-generation/editing slice covers fragment growing, scaffold hopping, and linker design. Side-chain decoration is reported separately as a budget-transfer stress test because it changes the output family from ligand-coordinate generation to narrow protein-local packing.

Our contributions are:
\begin{enumerate}[leftmargin=1.2em]
    \item \textbf{A quotient-lifted trust-region formulation.} We derive a metric-horizontal lift of quotient covectors and a closed-form sampler-relative correction whose norm is capped by the frozen sampler's own motion. For Gaussian reverse kernels, the active-budget update is the optimizer of a local KL-budgeted quotient-descent problem; for deterministic samplers, it is the corresponding kinetic trust-region rule. Deployed variants only instantiate where the quotient direction is evaluated.
    \item \textbf{Theory that matches the implementation questions.} We prove the horizontal lift, sampler-budgeted descent update, KL/kinetic budget interpretation, equivariance of the correction under isometric group actions, and product-budget split that defines a different family of motions than one-branch scalarization.
    \item \textbf{A practical quality--runtime evaluation.} Controlled tasks expose the dormant-gradient failure mode. Frozen \TargetDiff experiments show that deployed \qrg variants improve fragment and scaffold validity at modest overhead while preserving novelty and diversity in the matched multi-seed molecular slice. Linker design is near-neutral for \pcmain but improves for \cheapmain. Paired target-level analyses expose heterogeneity rather than hiding it, and a full side-chain sweep is included as a stress test showing task-dependent budget mismatch. Docking is reported as a corrected supporting diagnostic, not as an overclaimed headline.
\end{enumerate}

\section{Background}\label{sec:background}

\paragraph{Groups, invariants, and quotient coordinates.}
Let a group $G$ act on a data space $\Omega$ by $g\circ x$. A function $f:\Omega\rightarrow\mathbb{R}$ is $G$-invariant if $f(g\circ x)=f(x)$ for every $g\in G$, and a map $F:\Omega\rightarrow\Omega$ is $G$-equivariant if $F(g\circ x)=g\circ F(x)$. Molecular coordinates are typically considered up to rigid motions, and molecular graphs may have atom-index automorphisms. A quotient representation $q:\Omega\rightarrow\mathcal{Q}$ identifies states within the same equivalence class: $q(g\circ x)=q(x)$. Pairwise distances are the simplest example for rigid-motion invariance; pocket-relative or substructure-restricted distances give task-specific quotients.

\paragraph{Diffusion samplers and frozen-backbone guidance.}
A denoising diffusion model constructs noisy states $x_t$ and learns a reverse-time denoising update \citep{sohl2015nonequilibrium,song2019ncsn,ho2020ddpm,song2021score,nichol2021improved,kingma2021vdm,karras2022edm}. We write one sampler step abstractly as
\begin{equation}
    x_{t-1}=x_t+v_\theta(x_t,t),
\end{equation}
where $v_\theta$ includes the neural denoiser and the numerical update rule. Plug-and-play guidance modifies this step by adding a correction $u_t$ derived from an auxiliary objective, giving $x_{t-1}=x_t+v_\theta(x_t,t)+u_t$. Classifier guidance and classifier-free guidance popularized this idea in image generation \citep{dhariwal2021diffusion,ho2022classifierfree}; later methods use diffusion guidance for inverse problems, restoration, inpainting, and general differentiable objectives \citep{chung2023dps,kawar2022ddrm,lugmayr2022repaint,bansal2023universal}. Our setting differs in that the objective is often invariant while the correction must be delivered in ambient molecular coordinates.

\paragraph{Equivariant molecular diffusion.}
Equivariant neural networks encode symmetry by construction \citep{cohen2016group,kondor2018compact,thomas2018tensorfield,weiler2018steerable,anderson2019cormorant,fuchs2020se3transformer,finzi2020lieconv,satorras2021egnn,batzner2022nequip}; broader surveys frame these architectures as part of geometric deep learning \citep{bronstein2021geometric}. Molecular diffusion models use these ideas to generate 3D conformers, molecules, docking poses, linkers, and ligands in protein pockets \citep{gebauer2019gschnet,hoogeboom2022edm,xu2022geodiff,jing2022torsional,corso2023diffdock,guan2023targetdiff,igashov2024difflinker,schneuing2024diffsbdd}. Recent work also revisits symmetry-aware training itself: Orbit Diffusion treats data augmentation as a Monte Carlo gradient estimator and Rao-Blackwellizes over group orbits to reduce variance without extra neural function evaluations \citep{tong2025orbdiff}. Our paper is complementary. Rather than changing the training objective, we ask how to use quotient information at sampling time when the backbone is frozen.

\paragraph{Structure-based generation and lead optimization.}
Pocket-conditioned 3D molecular generation aims to produce valid molecules that satisfy geometric, chemical, and interaction constraints inside a protein binding pocket. Early models used 3D density grids \citep{ragoza2022ligan}; later models introduced equivariant graph sampling \citep{peng2022pocket2mol} and diffusion backbones \citep{guan2023targetdiff,schneuing2024diffsbdd,igashov2024difflinker,guan2024decompdiff}. Linker and fragment-editing settings also have task-specific graph and 3D generative models \citep{imrie2020delinker}. \CBGBench unifies de novo generation and lead-optimization tasks as completion of a protein-molecule binding graph, including linker design, fragment growing, scaffold hopping, and side-chain decoration \citep{lin2025cbgbench}. This multi-task view is important for our evaluation because a guidance method that only works on one custom subset is unlikely to be useful.

\paragraph{Evaluation pitfalls.}
Generation validity, novelty, diversity, and substructure consistency measure different aspects of molecular quality, and benchmark suites for molecular generation standardize related distributional and optimization metrics \citep{brown2019guacamol,polykovskiy2020moses}. QED and related cheminformatics scores are useful descriptors but should not be overread as evidence of drug-likeness \citep{bickerton2012qed}. Docking scores such as \Vina are useful diagnostics, but raw scores can be affected by molecular size, strain, coordinate-frame choices, and post-processing conventions \citep{wang2005pdbbind,trott2010vina,eberhardt2021vina}. Recent structure-based benchmarks and datasets emphasize that raw pocket-based 3D generators can produce invalid conformations and that local relaxation or dataset construction can substantially change apparent affinity estimates \citep{francoeur2020crossdocked,baillif2024genbench3d}. Our metric pipeline relies on standard cheminformatics toolkits for molecular parsing and conversion \citep{rdkit2026,oboyle2011openbabel}. We therefore separate the main generation claim from docking diagnostics and use corrected raw-pocket docking only as supporting evidence.

\section{Budgeted quotient-residual guidance}\label{sec:method}

We describe the method at the level of a frozen sampler. This keeps the formulation independent of a particular diffusion discretization and makes the runtime accounting explicit. The core object is a local, first-order quotient-lifted trust-region update; named variants only specify where the quotient direction is evaluated. \Cref{fig:overview} summarizes the deployed step in algorithm form, while the subsections below give the corresponding optimization problem and closed-form updates.

\subsection{Quotient differential and horizontal lift}

Let $x\in\Omega\subset\R^d$ denote ambient molecular coordinates and let $q:\Omega\rightarrow\mathcal{Q}\subset\R^m$ be a local quotient chart, such as pocket-relative distances, anchor distances, or substructure features. A task energy has the form
\begin{equation}
    E_{\rm quo}(x;y)=\ell(q(x),y),
\end{equation}
where $y$ is specified in quotient space. Write $J_x=Dq_x\in\R^{m\times d}$ for the quotient differential. Ambient motions in $\ker J_x$ are vertical: to first order they move the representative without moving the quotient features. A quotient covector $c_x=\nabla_z\ell(z,y)|_{z=q(x)}$ pulls back to the ambient covector $J_x^\top c_x$, but this pullback alone does not say how far a frozen sampler should move.

To separate quotient geometry from delivery scale, choose a positive definite sampler metric $M_x$ on ambient motions. In implementation $M_x$ is the Euclidean metric with optional lightweight feature normalization, but the metric notation makes the geometry explicit. The metric-horizontal lift of the quotient covector is
\begin{equation}
    h_x=M_x^{-1}J_x^\top c_x .
    \label{eq:horizontal_lift}
\end{equation}
It is horizontal because $\inner{h_x}{w}_{M_x}=0$ for every $w\in\ker J_x$, where $\inner{a}{b}_{M}=a^\top M b$. Thus \cref{eq:horizontal_lift} is a metric-horizontal ambient representative of the local quotient gradient; the normalized direction $-h_x/\norm{h_x}_{M_x}$ is the corresponding steepest quotient descent direction under the metric.

\begin{proposition}[Horizontal steepest lift]\label{prop:lift}
Let $J\in\R^{m\times d}$, $c\in\R^m$, and $M\succ0$. If $h=M^{-1}J^\top c\ne0$, then $-h/\norm{h}_M$ solves
\begin{equation}
    \min_{\norm{u}_M\le 1}\; \inner{c}{Ju} .
    \label{eq:horizontal_lift_problem}
\end{equation}
Moreover $h$ is $M$-orthogonal to $\ker J$.
\end{proposition}

This proposition is the quotient analogue of choosing the score direction in ambient diffusion: it says which ambient direction most efficiently changes the quotient objective under a local metric. It does not yet choose the physical amount of motion to deliver.

\subsection{Sampler-relative trust-region delivery}

At denoising step $t$, the frozen sampler proposes $v_t=v_\theta(x_t,t)$. We use the sampler's own step norm to define the trust radius
\begin{equation}
    B_t=\rho_t\norm{v_t}_{M_t},
\end{equation}
where $M_t=M_{\tilde{x}_t}$ and $\tilde{x}_t$ is the state where the quotient direction is evaluated. The delivered quotient correction solves the local problem
\begin{equation}
    \min_{u\in\R^d}\; \inner{c_t}{J_tu}
    \quad \emph{s.t.}\quad \norm{u}_{M_t}\le B_t .
    \label{eq:qrg_trust_problem}
\end{equation}
By \cref{prop:lift}, the active-budget solution is a sampler-scaled horizontal lift:
\begin{equation}
    u_t^{\rm quo}=-B_t\frac{h_t}{\norm{h_t}_{M_t}+\varepsilon},
    \qquad h_t=M_t^{-1}J_t^\top c_t .
    \label{eq:qrg_trust_update}
\end{equation}
The molecular implementation uses the budget-capped counterpart
\begin{equation}
    u_t^{\rm quo}=-\min\left\{1,\frac{B_t}{\norm{h_t}_{M_t}+\varepsilon}\right\}h_t,
    \label{eq:qrg_capped_update}
\end{equation}
which matches \cref{eq:qrg_trust_update} when the raw lift exceeds the budget and otherwise leaves the sub-budget lift unchanged. The key point is the separation: the quotient differential chooses a horizontal direction, while the frozen sampler bounds the delivery scale.

\begin{proposition}[Sampler-scaled quotient descent]\label{prop:trust}
Assume $h_t\ne0$ and ignore the numerical stabilizer. The active-budget update in \cref{eq:qrg_trust_update} is the unique optimizer of \cref{eq:qrg_trust_problem}. Its first-order quotient-energy change is
\begin{equation}
    \inner{c_t}{J_tu_t^{\rm quo}}
    =-\rho_t\norm{v_t}_{M_t}\norm{h_t}_{M_t} .
    \label{eq:first_order_descent}
\end{equation}
Thus the full-budget decrease is controlled by the sampler-relative budget, not by the raw quotient-gradient norm. The deployed capped update in \cref{eq:qrg_capped_update} obeys the same norm budget and coincides with this optimizer whenever the branch is budget active.
\end{proposition}

\paragraph{KL/kinetic interpretation.}
When a reverse step is written as a Gaussian reference kernel
\begin{equation}
    P_t(\mathrm{d}x_{t-1}\mid x_t)=\mathcal{N}\left(\mu_t,\tau_t M_t^{-1}\right),
    \qquad \mu_t=x_t+v_t,
\end{equation}
a mean-shifted control $u$ with the same covariance has conditional divergence
\begin{equation}
    \operatorname{KL}\left(P_t^u(\cdot\mid x_t)\,\Vert\,P_t(\cdot\mid x_t)\right)
    =\frac{1}{2\tau_t}\norm{u}_{M_t}^2 .
\end{equation}
Thus the metric trust region in \cref{eq:qrg_trust_problem} is also a local KL budget for stochastic reverse kernels, and a kinetic-energy budget in deterministic samplers, echoing the control-cost viewpoint in path-integral and linearly solvable control \citep{kappen2005path,todorov2009efficient}. Setting $B_t=\rho_t\norm{v_t}_{M_t}$ spends a fixed fraction of the frozen sampler's own step scale rather than an absolute ambient length.

\begin{proposition}[KL-budgeted quotient descent]\label{prop:kl_budget}
Assume the Gaussian reference kernel above, $h_t\ne0$, and ignore the numerical stabilizer. Among mean-shift controls satisfying
\begin{equation}
    \operatorname{KL}\left(P_t^u(\cdot\mid x_t)\,\Vert\,P_t(\cdot\mid x_t)\right)
    \le \kappa_t,
\end{equation}
the update that minimizes the linearized quotient loss $\inner{c_t}{J_tu}$ is
\begin{equation}
    u_t^\star=-\sqrt{2\tau_t\kappa_t}\frac{h_t}{\norm{h_t}_{M_t}}.
\end{equation}
The QRG radius $B_t=\rho_t\norm{v_t}_{M_t}$ corresponds to the conditional budget $\kappa_t=\rho_t^2\norm{v_t}_{M_t}^2/(2\tau_t)$.
\end{proposition}

\subsection{Section-residual product budget}

Some molecular tasks require preserving a concrete representative as well as improving quotient features. We therefore use a product trust region with independently budgeted section and residual directions. The section branch uses $h_t^{\rm sec}=M_t^{-1}\nabla_xE_{\rm sec}(\tilde{x}_t)$ to stabilize anchors or preserved atoms. The residual branch uses the quotient-horizontal lift $h_t^{\rm res}=M_t^{-1}J_t^\top c_t$ from \cref{eq:horizontal_lift}. The delivered update is
\begin{equation}
    u_t = -\min\left\{1,\frac{\rho_s\norm{v_t}_{M_t}}{\norm{h_t^{\rm sec}}_{M_t}+\varepsilon}\right\}h_t^{\rm sec}
        -\min\left\{1,\frac{\rho_r\norm{v_t}_{M_t}}{\norm{h_t^{\rm res}}_{M_t}+\varepsilon}\right\}h_t^{\rm res}.
    \label{eq:split_update}
\end{equation}
The endpoint $\rho_r=0$ is \sectiononly. Active variants allocate a small residual budget $\rho_r>0$. This is not a single mixed objective followed by one projection; it is a product-budget allocation over two separately capped directions, one for representative preservation and one for quotient-residual descent.

\begin{proposition}[Product-budget split is not one-branch scalarization]\label{prop:split}
Let $a,b\in\R^d$ be non-collinear and let $B_a,B_b>0$. The split update $-B_a a/\norm{a}-B_b b/\norm{b}$ cannot in general be written as $-B(a+\lambda b)/\norm{a+\lambda b}$ for a fixed scalar $\lambda$ and budget $B$ that is independent of the branch allocation. Thus product-budget residual delivery defines a different family of delivered motions than one-branch scalarization.
\end{proposition}

\paragraph{Relation to normalized guidance.}
Normalized plug-and-play guidance divides by a gradient norm and still requires an absolute step length in ambient coordinates. \qrg instead uses the frozen sampler's own motion as a branch budget; the deployed setting clips each lifted branch to that budget, while the full-budget normalized rule is a useful limiting case. The completed controls compare the deployed residual family against no guidance, section-only delivery, shuffled residuals, and rollout-heavy teacher guidance; we do not claim these controls exhaust every possible scalarized alternative.

\subsection{Instantiating the quotient direction}

The trust-region mechanism does not require rollout. It only requires a state $\tilde{x}_t$ at which to evaluate $J_t$ and $c_t$. This separates the method from its instantiations: all variants below use the same quotient-lifted delivery rule and differ only in how cheaply they obtain the quotient features.

\paragraph{Predicted-next quotient-residual guidance.}
\pcmain reuses the frozen sampler's own one-step prediction, $\tilde{x}_t=x_t+v_t$, and applies the residual correction every two denoising steps in the reported configuration. This variant has the cleanest deployment story because it does not ask the denoiser for a separate rollout trajectory.

\paragraph{Local-radius quotient surrogate.}
\cheapmain uses the same trust-region delivery but computes the quotient residual from a local-radius feature subset rather than all dense pairwise terms. This variant is the stronger practical point in the completed runs: it pays moderate feature-computation overhead but avoids the many denoiser calls of rollout lookahead. A development ablation in \cref{tab:local_radius_development} compares this surrogate with sampled-pair alternatives.

\paragraph{Shared-rollout teacher.}
\teacher evaluates the quotient direction after one explicit short rollout and is useful as a teacher or upper-bound diagnostic. It is not the deployed method. Reporting it separately prevents the paper from hiding expensive inference behind a single method name.

\subsection{Equivariance of the lift and delivery}

Let $G$ be an isometry group acting linearly by $R_g$ on ambient coordinates. The quotient chart is invariant if $q(R_gx)=q(x)$, and the metric is equivariant if $M_{R_gx}=R_gM_xR_g^\top$.

\begin{proposition}[Equivariant quotient-lifted correction]\label{prop:equivariance}
Assume $q(R_gx)=q(x)$, $M_{R_gx}=R_gM_xR_g^\top$, and the frozen sampler is equivariant: $v_\theta(R_gx,t)=R_gv_\theta(x,t)$. Then the horizontal lift and delivered residual correction satisfy
\begin{equation}
    h_{R_gx}=R_gh_x,
    \qquad
    u_t^{\rm quo}(R_gx_t)=R_gu_t^{\rm quo}(x_t).
\end{equation}
\end{proposition}

This proposition does not claim that the frozen backbone becomes perfectly equivariant; it states that the guidance correction itself respects the quotient symmetry whenever its inputs do.

\subsection{What the method is not}

\qrg is not a new trained backbone and does not claim to replace equivariant architectures or symmetry-aware losses. It is also not an unbounded external optimizer wrapped around a sampler. The method is intentionally small: it spends a bounded amount of motion in a quotient-residual direction and measures the resulting quality--runtime trade-off. This is why all experiments report wall-clock runtime and include controls such as \sectiononly, \sham, and the rollout teacher.
Proofs and implementation details are provided in \cref{app:proofs,app:expdetails}.

\FloatBarrier
\section{Experimental setup and results}\label{sec:experiments}

We organize the experiments around three questions. First, does sampler-relative delivery activate quotient information that raw local gradients leave dormant? Second, on frozen pocket-conditioned molecular samplers, does the deployed method improve the quality--runtime trade-off without collapsing novelty or diversity? Third, where does the same budget fail? The main ligand-generation/editing evaluation covers fragment growing, scaffold hopping, and linker design. Side-chain decoration is reported separately as a budget-transfer stress test.

\paragraph{Evidence map.}
We keep the evidence types separate. Official seed-0 generation sweeps carry the main validity, runtime, and control-budget claim. Paired target-level statistics test whether aggregate deltas are driven by broad target-level movement or by heterogeneous pockets. The matched multi-seed molecular slice supports novelty and diversity preservation. Corrected \Vina is a supporting raw-pocket diagnostic, not a binding-affinity claim. Additional result views, compute accounting, reproducibility notes, and related-work context are in \cref{app:additional,app:impact_compute,app:repro,app:related}.

\subsection{Controlled quotient tasks: does trust-region delivery activate the signal?}

\paragraph{Setup.}
We first use three controlled tasks where the correct objective is defined in quotient space: a 2D Gaussian task with nuisance rotations, an orbit-point matching task, and a toy 3D molecule task using invariant distance templates. Each task uses the same denoising schedule for the base sampler, dormant local quotient guidance, and budgeted residual guidance. We measure target success and a task-specific quotient error. These toys are not intended as molecular benchmarks; they isolate whether the proposed delivery rule fixes dormant local gradients.

\begin{figure}[t]
    \centering
    \includegraphics[width=\linewidth]{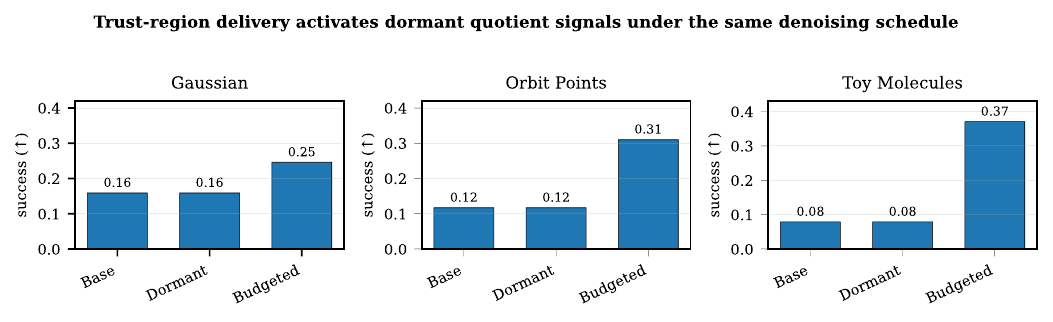}
    \caption{Controlled quotient tasks. Dormant local guidance is almost identical to the base sampler, while trust-region delivery makes the same quotient signal active under the same denoising schedule.}
    \label{fig:toy_activation}
\end{figure}

\begin{table}[t]
    \caption{Toy-task summary. The key error is quotient MMD for Gaussian 2D and invariant template error for the geometry tasks. All rows within a task use the same denoising budget.}
    \label{tab:toy_summary}
    \centering
    \small
    \begin{tabular}{llccc}
\toprule
Task & Method & Success $\uparrow$ & Key error $\downarrow$ & NFE \\
\midrule
\multirow{3}{*}{Gaussian 2D}
 & \base & 0.159 & 0.528 & 64 \\
 & Dormant local & 0.159 & 0.529 & 64 \\
 & Budgeted residual & \textbf{0.246} & \textbf{0.451} & 64 \\
\midrule
\multirow{3}{*}{Orbit points}
 & \base & 0.117 & 0.406 & 72 \\
 & Dormant local & 0.117 & 0.406 & 72 \\
 & Budgeted residual & \textbf{0.310} & \textbf{0.299} & 72 \\
\midrule
\multirow{3}{*}{Toy molecules}
 & \base & 0.079 & 0.514 & 80 \\
 & Dormant local & 0.079 & 0.514 & 80 \\
 & Budgeted residual & \textbf{0.371} & \textbf{0.356} & 80 \\
\bottomrule
\end{tabular}

\end{table}

\paragraph{Result.}
\Cref{fig:toy_activation,tab:toy_summary} show the intended mechanism. Raw local quotient guidance remains dormant: success and quotient error are essentially unchanged from the base sampler. Budgeted residual guidance increases success on all tasks, from $0.159$ to $0.246$ on Gaussian 2D, from $0.117$ to $0.310$ on orbit points, and from $0.079$ to $0.371$ on toy molecules. This supports the central design choice before moving to molecular generation: the quotient direction needs sampler-relative delivery, not just a larger list of rollout heuristics.

\subsection{Budgeted inference on frozen pocket-conditioned samplers}

\paragraph{Setup.}
We next evaluate frozen \TargetDiff inside a \CBGBench-style lead-optimization protocol. Each target generates the same number of samples under the base sampler and guided variants, and no model weights are updated. The two deployed variants instantiate the same trust-region update: \pcmain reuses the predicted next state and applies residual guidance every two steps, while \cheapmain uses a local-radius quotient surrogate. The official seed-0 table reports completed aggregate rows for the ligand-generation/editing tasks, and the novelty/diversity columns summarize the matched multi-seed molecular slice for the same task and method.

\begin{table*}[t]
    \caption{Budgeted inference on frozen ligand-generation/editing tasks. Validity, runtime, and control/base ratio are official seed-0 aggregates; novelty and diversity are from the matched multi-seed molecular slice for the same task and method. Fragment and scaffold improve for both deployed variants. Linker design is near-neutral for \pcmain but improves for \cheapmain, showing that the quality--runtime trade-off is task-dependent rather than uniformly monotone.}
    \label{tab:molecular_main}
    \centering
    \small
    \resizebox{\textwidth}{!}{\begin{tabular}{llcccccccc}
\toprule
Task & Method & Official parts & Validity $\uparrow$ & $\Delta$ val. & sec/sample & Control/base & Novelty $\uparrow$ & Diversity $\uparrow$ \\
\midrule
\multirow{3}{*}{Fragment}
 & \base & 5/5 & 0.815 & -- & 16.68 & 0.000 & 1.000 & 0.898 \\
 & \pcmain & 5/5 & 0.829 & +0.014 & 18.28 & 0.192 & 1.000 & 0.899 \\
 & \cheapmain & 5/5 & \textbf{0.864} & \textbf{+0.049} & 19.91 & 0.265 & 1.000 & 0.896 \\
\midrule
\multirow{3}{*}{Scaffold}
 & \base & 5/5 & 0.664 & -- & 17.99 & 0.000 & 1.000 & 0.913 \\
 & \pcmain & 5/5 & 0.683 & +0.019 & 20.36 & 0.194 & 1.000 & 0.911 \\
 & \cheapmain & 5/5 & \textbf{0.707} & \textbf{+0.043} & 20.18 & 0.268 & 1.000 & 0.912 \\
\midrule
\multirow{3}{*}{Linker}
 & \base & 5/5 & 0.681 & -- & 18.82 & 0.000 & 1.000 & 0.888 \\
 & \pcmain & 5/5 & 0.680 & -0.001 & 19.72 & 0.183 & 1.000 & 0.892 \\
 & \cheapmain & 5/5 & \textbf{0.712} & \textbf{+0.031} & 21.38 & 0.251 & 1.000 & 0.891 \\
\bottomrule
\end{tabular}
}
\end{table*}

\paragraph{Result.}
\Cref{tab:molecular_main} gives the official ligand-generation/editing picture. On fragment growing, \pcmain improves validity from $0.815$ to $0.829$, and \cheapmain improves it to $0.864$. On scaffold hopping, \pcmain improves validity from $0.664$ to $0.683$, and \cheapmain improves it to $0.707$. The control/base ratios remain in the intended budgeted regime, around $0.19$ for \pcmain and $0.27$ for \cheapmain. Linker design is harder for \pcmain, which is near the base sampler at $0.680$ versus $0.681$, but \cheapmain improves the completed aggregate to $0.712$. Paired target-level analyses in \cref{tab:paired_validity} show the strongest support on fragment for \cheapmain (mean $\Delta{=}+0.049$, interval $[0.019,0.078]$), a positive scaffold aggregate with target-level heterogeneity, and a positive linker aggregate for \cheapmain with a target-level interval that crosses zero. Across the matched multi-seed molecular slice, novelty remains $1.000$ and diversity stays close to the base sampler. We therefore phrase the result as a bounded quality--runtime trade-off rather than a uniform improvement theorem.

\begin{figure}[t]
    \centering
    \includegraphics[width=\linewidth]{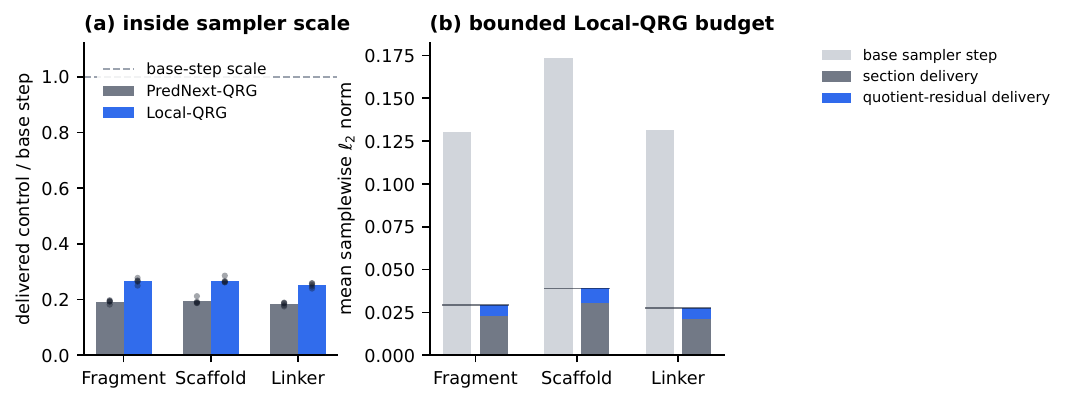}
    \caption{Official-run norm accounting for the trust-region mechanism. Left: delivered control stays well below the frozen sampler's own step scale across official seed-0 ligand-generation/editing shards. Right: for \cheapmain, the delivered section and quotient-residual components spend a small bounded budget relative to the base sampler motion. This is a kinetic-budget diagnostic, not a measured path KL: it supports the mechanism claim that \qrg delivers quotient information at sampler scale rather than applying an unbounded external optimizer.}
    \label{fig:mechanism_norms}
\end{figure}

\subsection{Anti-reparameterization controls}

\paragraph{Setup.}
A guidance method should not be evaluated only by its best score if that score hides a large sampler cost or if any arbitrary second branch would work. We therefore include controls that target the main alternative explanations available in the completed runs. \sectiononly sets the residual budget to zero; \sham uses a shuffled residual signal; \teacher uses one shared rollout and is treated as a teacher/oracle rather than the deployed method. The main empirical comparison here is the deployed residual family against no-guidance, section-only, shuffled-residual, and rollout-teacher controls.

\paragraph{Result.}
\Cref{tab:ablation_frontier} supports three conclusions. First, the deployed points occupy a practical runtime regime rather than a many-rollout regime. Second, expensive rollout is not necessary for the observed validity gains; the rollout teacher is useful for analysis but is not the deployed method. Third, the fragment controls show that structured residual guidance is not simply ``any second branch'': \sham is weaker than \cheapmain, though close enough to \pcmain that we avoid claiming this control alone rules out every possible scalarized-guidance explanation.

\subsection{Side-chain decoration as a budget stress test}

\paragraph{Setup.}
Side-chain decoration is an official benchmark task, but it is not the same geometric family as ligand fragment growing, scaffold hopping, or linker design. The output is constrained by narrow local packing and rotamer-like geometry. We therefore use the completed full side-chain sweep as a stress test for the budget, not as part of the ligand-generation win table.

\paragraph{Result.}
On the full side-chain sweep, unguided validity is $0.873$, while \pcmain, \cheapmain, and \sectiononly obtain $0.848$, $0.837$, and $0.828$. Paired target statistics are negative for the default guided settings. Reducing the budget and applying guidance sparsely nearly rescues performance: the sparse \cheapmain variant reaches $0.869$, with a paired confidence interval crossing zero. This is a useful stress test for budget transfer. It shows that the same quotient-residual mechanism can diagnose when the delivered budget is wrong: ligand-generation budgets improve broader ligand-editing tasks but over-control narrow side-chain packing. The final claim is therefore a task-dependent quality--runtime trade-off, not a universal monotonic improvement rule.

\paragraph{Corrected docking diagnostic.}
Because raw docking is confounded by coordinate frames, geometry, and molecular size, we report fixed raw-pocket \Vina only as supporting evidence. \Cref{tab:vina_summary} shows the clearest signal on scaffold hopping for \cheapmain (paired mean $-0.285$, interval $[-0.397,-0.178]$), with fragment mild and linker mixed; the main claim remains validity and quality--runtime behavior.

\section{Conclusion and limitations}\label{sec:conclusion}

\qrg turns quotient-residual directions into bounded sampler-time corrections for frozen molecular diffusion. The completed evidence supports a practical but scoped claim: quotient-aware residual delivery improves the quality--runtime trade-off on the main frozen-backbone ligand-generation/editing tasks, with the clearest gains on fragment growing and scaffold hopping and a useful \cheapmain gain on linker design. The controlled tasks isolate the mechanism before moving to molecules, the official seed-0 sweeps provide the main validity/runtime accounting, and the matched molecular slice shows that the gains do not come from an obvious novelty or diversity collapse. The boundaries are equally important: paired target statistics reveal heterogeneity, docking remains a supporting mixed diagnostic, and side-chain decoration exposes a budget-transfer failure mode rather than another headline win.

\bibliographystyle{plainnat}
\bibliography{refs}

\clearpage
\appendix
\section{Expanded related work}\label{app:related}

\paragraph{Equivariant architectures.}
Group-equivariant convolutional networks formalized convolutional symmetry beyond translations \citep{cohen2016group}, and compact-group equivariance results clarified when convolutional structure is forced by symmetry \citep{kondor2018compact}. Tensor Field Networks, 3D steerable CNNs, SE(3)-Transformers, and LieConv introduced rotation-aware or Lie-group-aware message passing for 3D geometric data \citep{thomas2018tensorfield,weiler2018steerable,fuchs2020se3transformer,finzi2020lieconv}; EGNNs gave a simpler coordinate-update architecture for Euclidean equivariance \citep{satorras2021egnn}. Molecular and atomistic models such as Cormorant and NequIP show the value of covariant features in molecular property prediction and interatomic potentials \citep{anderson2019cormorant,batzner2022nequip}. These models are powerful but can be more complex or expensive than non-equivariant alternatives, motivating complementary approaches that preserve or exploit symmetry without changing the entire architecture.

\paragraph{Equivariant diffusion and flows.}
Equivariant normalizing flows and flow matching models learn densities or transports under known group actions \citep{kohler2020equivariant,klein2024equivariantflow}. Symmetry-adapted 3D point-set generation and equivariant diffusion for 3D molecules showed how to generate atom types and coordinates with Euclidean symmetries \citep{gebauer2019gschnet,hoogeboom2022edm}; GeoDiff and Torsional Diffusion focus on conformer generation \citep{xu2022geodiff,jing2022torsional}. In proteins, SE(3) diffusion and flow matching have been used for backbone generation \citep{yim2023se3diffusion,bose2024foldflow}. Our method is not an architectural contribution in this line; it is an inference-time correction applied to a frozen model.

\paragraph{Rao-Blackwellization and orbit averaging.}
Orbit Diffusion interprets data augmentation as a Monte Carlo gradient estimator of a symmetrized loss and reduces variance by estimating an orbit-conditional expectation \citep{tong2025orbdiff}. Its practical implementation requires only one neural forward/backward pass per training example, because the additional work occurs in the target construction rather than through repeated model evaluations. This observation strongly informs our runtime goal: symmetry-aware or quotient-aware improvements are most convincing when they do not hide large neural evaluation overhead.

\paragraph{KL-budgeted control.}
Path-integral and linearly solvable optimal-control formulations make relative-entropy or quadratic control effort central to the control objective \citep{kappen2005path,todorov2009efficient}. The Schr{\"o}dinger problem gives a broader path-space entropy-regularized transport viewpoint \citep{leonard2014schrodinger}, and diffusion Schr{\"o}dinger bridges bring this path-space KL viewpoint into score-based generative modeling \citep{debortoli2021dsb}. QRG uses only the local one-step consequence of this viewpoint: for a Gaussian reverse kernel, a mean shift has a closed-form conditional KL, so the sampler-relative radius can be read as a one-step kinetic or KL budget rather than as an arbitrary ambient learning rate.

\paragraph{SBDD and benchmark design.}
Structure-based generation has progressed from grid-based latent models \citep{ragoza2022ligan} to pocket-conditioned graph samplers \citep{peng2022pocket2mol} and 3D diffusion models \citep{guan2023targetdiff,schneuing2024diffsbdd,guan2024decompdiff}. Task-specific models address docking, linker design, scaffold-oriented generation, and molecular property prediction with 3D pretraining \citep{corso2023diffdock,imrie2020delinker,igashov2024difflinker,stark2022infomax}. \CBGBench unifies these tasks through binding-graph completion and emphasizes interaction, chemical, geometric, and substructure metrics \citep{lin2025cbgbench}. \GenBench and CrossDocked-style datasets highlight the importance of conformation validity, dataset construction, and the risk of over-interpreting raw docking scores from invalid geometries \citep{francoeur2020crossdocked,baillif2024genbench3d}. General molecular-generation benchmark suites provide complementary distributional metrics such as validity, novelty, and diversity \citep{brown2019guacamol,polykovskiy2020moses}. These benchmarks shape our claim boundary.

\section{Proofs}\label{app:proofs}

\subsection{Proof of \cref{prop:lift}}
Let $h=M^{-1}J^\top c$. For any feasible $u$ in \cref{eq:horizontal_lift_problem},
\begin{equation}
    \inner{c}{Ju}=\inner{J^\top c}{u}=\inner{h}{u}_{M}
    \ge -\norm{h}_{M}\norm{u}_{M}
    \ge -\norm{h}_{M} .
\end{equation}
The lower bound is attained by $u^\star=-h/\norm{h}_{M}$, proving optimality. If $w\in\ker J$, then
\begin{equation}
    \inner{h}{w}_{M}=h^\top Mw=c^\top Jw=0,
\end{equation}
so $h$ is $M$-orthogonal to the vertical subspace $\ker J$.

\subsection{Proof of \cref{prop:trust}}
The objective in \cref{eq:qrg_trust_problem} can be written as $\inner{h_t}{u}_{M_t}$, where $h_t=M_t^{-1}J_t^\top c_t$. For any feasible $u$,
\begin{equation}
    \inner{h_t}{u}_{M_t}
    \ge -\norm{h_t}_{M_t}\norm{u}_{M_t}
    \ge -B_t\norm{h_t}_{M_t}.
\end{equation}
The lower bound is attained uniquely, for $h_t\ne0$, at $u_t^{\rm quo}=-B_t h_t/\norm{h_t}_{M_t}$. Substituting $B_t=\rho_t\norm{v_t}_{M_t}$ gives \cref{eq:qrg_trust_update}. The first-order quotient-energy change is
\begin{equation}
    \inner{c_t}{J_tu_t^{\rm quo}}
    =\inner{h_t}{u_t^{\rm quo}}_{M_t}
    =-\rho_t\norm{v_t}_{M_t}\norm{h_t}_{M_t}.
\end{equation}

\subsection{Proof of \cref{prop:kl_budget}}
For Gaussian kernels with the same covariance, the conditional KL divergence is the quadratic mean-shift cost
\begin{equation}
    \operatorname{KL}\left(P_t^u(\cdot\mid x_t)\,\Vert\,P_t(\cdot\mid x_t)\right)
    =\frac{1}{2}u^\top(\tau_tM_t^{-1})^{-1}u
    =\frac{1}{2\tau_t}\norm{u}_{M_t}^2 .
\end{equation}
The constraint $\operatorname{KL}(P_t^u\Vert P_t)\le\kappa_t$ is therefore equivalent to
\begin{equation}
    \norm{u}_{M_t}\le\sqrt{2\tau_t\kappa_t} .
\end{equation}
The linearized quotient loss satisfies
\begin{equation}
    \inner{c_t}{J_tu}=\inner{J_t^\top c_t}{u}=\inner{h_t}{u}_{M_t} .
\end{equation}
By Cauchy--Schwarz, every feasible $u$ obeys
\begin{equation}
    \inner{h_t}{u}_{M_t}
    \ge -\norm{h_t}_{M_t}\norm{u}_{M_t}
    \ge -\sqrt{2\tau_t\kappa_t}\norm{h_t}_{M_t} .
\end{equation}
The lower bound is attained uniquely, for $h_t\ne0$, by
\begin{equation}
    u_t^\star=-\sqrt{2\tau_t\kappa_t}\frac{h_t}{\norm{h_t}_{M_t}} .
\end{equation}
Choosing $\kappa_t=\rho_t^2\norm{v_t}_{M_t}^2/(2\tau_t)$ gives $\norm{u_t^\star}_{M_t}=\rho_t\norm{v_t}_{M_t}$, which is the QRG trust radius. For deterministic reverse samplers, the same calculation is read as the zero-noise kinetic trust-region analogue rather than as a measured path-space KL.

\subsection{Proof of \cref{prop:split}}
Let $a,b$ be non-collinear. The split family with independent branch budgets contains vectors
\begin{equation}
    s(B_a,B_b)=-B_a\frac{a}{\norm{a}}-B_b\frac{b}{\norm{b}}, \qquad B_a,B_b>0 .
\end{equation}
Changing the ratio $B_b/B_a$ changes the direction of $s(B_a,B_b)$ continuously in the cone spanned by $a$ and $b$. A one-branch scalarized delivery with fixed scalar $\lambda$ has direction
\begin{equation}
    m(\lambda)=-\frac{a+\lambda b}{\norm{a+\lambda b}},
\end{equation}
independent of the total budget $B$. Therefore no fixed $\lambda$ can recover the split directions for more than one branch-allocation ratio unless $a$ and $b$ are collinear. This proves that split delivery is not, in general, a scalar rescaling of a one-branch scalarized gradient.

\subsection{Proof of \cref{prop:equivariance}}
Because $q(R_gx)=q(x)$, differentiating gives
\begin{equation}
    J_{R_gx}R_g=J_x,
    \qquad\text{so}\qquad
    J_{R_gx}^\top=R_gJ_x^\top .
\end{equation}
The quotient covector $c_x=\nabla_z\ell(q(x),y)$ is unchanged because $q(R_gx)=q(x)$. Using $M_{R_gx}=R_gM_xR_g^\top$,
\begin{equation}
    h_{R_gx}=M_{R_gx}^{-1}J_{R_gx}^\top c_x
    =(R_gM_xR_g^\top)^{-1}R_gJ_x^\top c_x
    =R_gM_x^{-1}J_x^\top c_x
    =R_gh_x .
\end{equation}
The metric norm is preserved: $\norm{R_gw}_{M_{R_gx}}=\norm{w}_{M_x}$ for every $w$. The sampler equivariance assumption gives $v_\theta(R_gx,t)=R_gv_\theta(x,t)$, so the trust radius is also preserved. Substituting these identities into \cref{eq:qrg_trust_update} yields $u_t^{\rm quo}(R_gx_t)=R_gu_t^{\rm quo}(x_t)$.

\subsection{Why the trust-region update addresses dormancy}
Dormancy is not the absence of a useful direction; it is the mismatch between the raw quotient-gradient scale and the sampler's step scale. A raw update $-\eta h_t$ has delivered norm $\eta\norm{h_t}_{M_t}$, which can be arbitrarily small even when $h_t/\norm{h_t}_{M_t}$ is the horizontal steepest quotient direction. The trust-region update has delivered norm approximately $\rho_t\norm{v_t}_{M_t}$ for any nonzero lift. Thus the amount of motion is controlled by the sampler-relative budget rather than by the raw quotient-gradient magnitude. The toy tasks in \cref{sec:experiments} test exactly this mechanism.

\section{Experimental details}\label{app:expdetails}

\subsection{Toy quotient tasks}
The toy tasks are designed to stress quotient-aware guidance without introducing molecular engineering confounders. Each task uses the same frozen base denoising schedule for the base sampler, dormant local quotient guidance, and budgeted residual guidance. The Gaussian task evaluates quotient MMD. The orbit-point task and toy-molecule task evaluate invariant template errors. The key comparison is not absolute performance, but whether a direction that is inert under raw local delivery becomes active under trust-region delivery.

\subsection{Molecular generation protocol}
All molecular experiments use the \TargetDiff backbone and a \CBGBench-style lead-optimization protocol. The official seed-0 sweeps provide the full aggregate validity, runtime, and control-budget rows. The matched multi-seed molecular slice uses the same target/sample contract where available and supports novelty, diversity, development-ablation, sham-control, and teacher-diagnostic summaries. For each task and seed, the base sampler and guided variants generate matched samples under the same target set and sample contract. We report validity, uniqueness, QED, diversity, novelty, seconds per sample, and control/base ratio. Runtime values are fair wall-clock estimates after removing contaminated co-run measurements. The code supplement provides the QRG runner and summarization scripts; raw benchmark assets, checkpoints, generated molecules, logs, and precomputed result tables are not redistributed.

\subsection{Partial-result policy}
Main-text quantitative claims use completed aggregate rows only. A row is considered complete when the planned shards for that task--method pair have written \texttt{run\_summary.json} and have been included in the aggregate script. Rows with fewer completed parts are labeled by their completed-part count and are interpreted only as bounded aggregate evidence, not as a hidden full-sweep claim.

\subsection{Variant definitions}
\begin{itemize}[leftmargin=1.2em]
    \item \base: frozen sampler with no additional guidance.
    \item \sectiononly: section branch enabled, residual budget zero.
    \item \pcmain: predicted-next quotient-residual correction, every two denoising steps in the reported configuration.
    \item \cheapmain: local-radius quotient surrogate with trust-region delivery.
    \item \teacher: one shared rollout before evaluating the residual direction; reported only as a teacher/oracle point.
    \item \sham: residual branch with a shuffled or otherwise mismatched residual signal; tests whether any second branch can produce the gain.
\end{itemize}

\subsection{Docking diagnostic}
We use fixed raw-pocket boxes and target-balanced paired comparisons. Each paired row compares a guided variant to the base sampler on the same target. The statistic is the target-level mean difference in raw \Vina score; negative values are better. This diagnostic is intentionally separated from the main validity claim because raw docking can be confounded by molecular size, strain, and conformation validity.

\section{Additional completed results}\label{app:additional}

\subsection{Frontier figure}
\begin{figure}[h]
    \centering
    \includegraphics[width=0.96\linewidth]{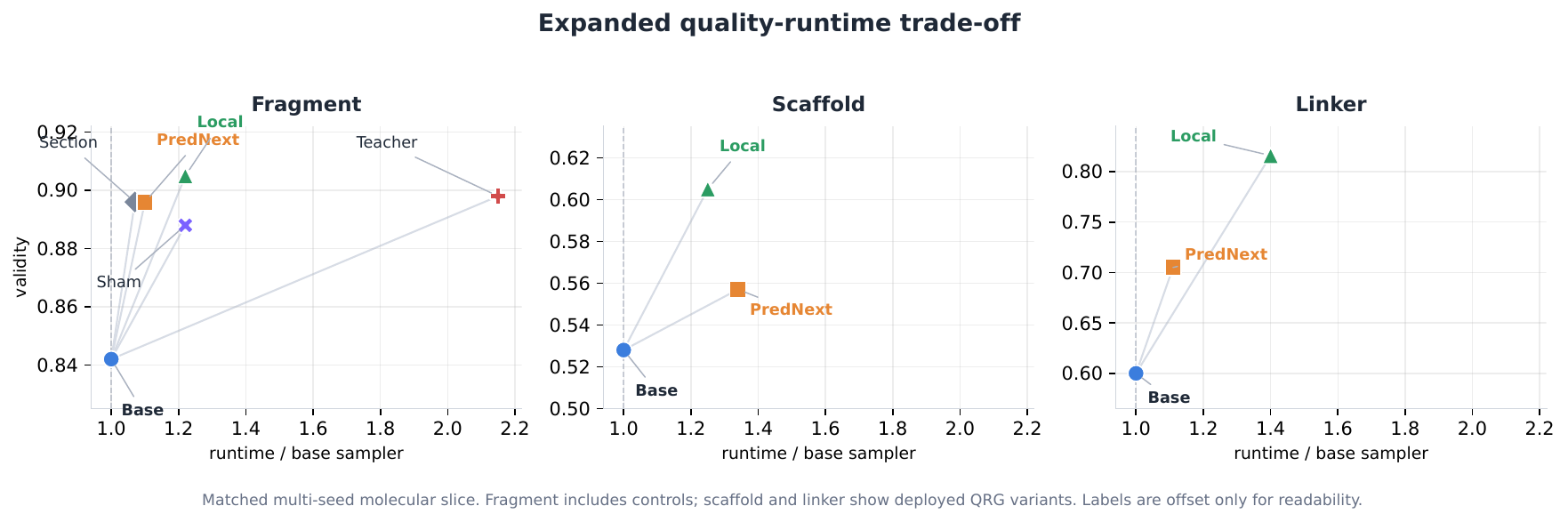}
    \caption{Expanded quality--runtime trade-off from the matched multi-seed molecular slice. Fragment has the richest ablation set; scaffold and linker include base, \pcmain, and \cheapmain. This figure complements the official seed-0 aggregate by showing novelty and diversity preservation across matched seeds.}
    \label{fig:frontier_expanded}
\end{figure}

\subsection{Generation metrics}
The detailed generation summaries record per-seed validity, uniqueness, QED, diversity, seconds per sample, and control/base ratio. The main text reports the metrics most relevant to the claim boundary: validity, runtime, novelty, and diversity. The code supplement contains the scripts used to recompute these summaries from generated outputs rather than precomputed result tables.

\subsection{Development ablation for the cheap surrogate}
\begin{table}[h]
    \caption{Development-slice fragment seed-0 ablation for cheap quotient surrogates. The local-radius surrogate is not introduced as an arbitrary extra branch: on this slice it outperforms sampled-pair residual surrogates at comparable runtime while preserving diversity.}
    \label{tab:local_radius_development}
    \centering
    \small
    \begin{tabular}{lcccc}
\toprule
Cheap surrogate & Validity $\uparrow$ & Unique & Diversity & sec/sample \\
\midrule
Sampled pairs, 16 & 0.888 & 854 & 0.896 & 17.84 \\
Sampled pairs, 128 & 0.886 & 856 & 0.897 & 18.15 \\
Local radius & \textbf{0.902} & \textbf{869} & 0.896 & 16.88 \\
\bottomrule
\end{tabular}

\end{table}

The local-radius variant was selected as the practical cheap surrogate before the official full sweeps. This ablation is intentionally scoped as development evidence rather than an additional official benchmark claim.

\subsection{Paired target-level validity}
\begin{table}[h]
    \caption{Paired official seed-0 target-level validity deltas. Each row compares a guided method to the unguided sampler on the same targets; bootstrap intervals resample targets. These statistics support the aggregate table while showing target heterogeneity.}
    \label{tab:paired_validity}
    \centering
    \small
    \begin{tabular}{llccclc}
\toprule
Task & Method & $n$ & Mean $\Delta$ & Median $\Delta$ & 95\% bootstrap CI & Wins/losses/ties \\
\midrule
Fragment & \pcmain & 61 & +0.013 & +0.010 & $[-0.006, 0.032]$ & 32/27/2 \\
Fragment & \cheapmain & 61 & \textbf{+0.049} & +0.030 & $[0.019, 0.078]$ & 42/14/5 \\
Scaffold & \pcmain & 64 & +0.019 & +0.040 & $[-0.029, 0.061]$ & 44/17/3 \\
Scaffold & \cheapmain & 64 & \textbf{+0.043} & +0.060 & $[-0.016, 0.099]$ & 43/19/2 \\
Linker & \pcmain & 43 & -0.001 & +0.020 & $[-0.037, 0.033]$ & 22/19/2 \\
Linker & \cheapmain & \textbf{43} & \textbf{+0.030} & +0.040 & $[-0.012, 0.073]$ & 25/14/4 \\
\bottomrule
\end{tabular}

\end{table}

The paired analysis matches the aggregate claim boundary. \cheapmain has the clearest fragment signal, with mean target-level validity delta $+0.049$ and bootstrap interval $[0.019,0.078]$. Scaffold has positive mean deltas for both deployed variants and positive win/loss counts, but wider target-level intervals. Linker is near-neutral for \pcmain and positive in aggregate for \cheapmain, with a mean paired delta of $+0.030$ and a bootstrap interval crossing zero.

\subsection{Target-level heterogeneity}
\begin{table}[h]
    \caption{Distribution of per-target official seed-0 validity deltas versus the unguided sampler. Each target contributes one target-level validity difference for the same task and method. The quantiles show that aggregate gains coexist with heterogeneous pockets, including a small number of large negative and large positive target-level shifts.}
    \label{tab:target_delta_distribution}
    \centering
    \scriptsize
    \setlength{\tabcolsep}{3.5pt}
    \begin{tabular}{llrrrrrrrr}
\toprule
Task & Method & $n$ & Mean $\Delta$ & Median & P10 & P25 & P75 & P90 & Min/Max \\
\midrule
Fragment & \pcmain & 61 & +0.013 & +0.010 & -0.060 & -0.020 & +0.050 & +0.100 & -0.160/+0.210 \\
Fragment & \cheapmain & 61 & +0.049 & +0.030 & -0.060 & +0.000 & +0.100 & +0.190 & -0.280/+0.400 \\
Scaffold & \pcmain & 64 & +0.019 & +0.040 & -0.097 & -0.010 & +0.103 & +0.167 & -0.860/+0.330 \\
Scaffold & \cheapmain & 64 & +0.043 & +0.060 & -0.246 & -0.023 & +0.190 & +0.271 & -0.860/+0.520 \\
Linker & \pcmain & 43 & -0.001 & +0.020 & -0.158 & -0.035 & +0.060 & +0.130 & -0.340/+0.270 \\
Linker & \cheapmain & 43 & +0.030 & +0.040 & -0.148 & -0.050 & +0.130 & +0.156 & -0.260/+0.440 \\
\bottomrule
\end{tabular}

\end{table}

Rather than listing all 504 task--method--target rows in the PDF, \cref{tab:target_delta_distribution} summarizes the complete per-target delta file. This makes the amount of target-level evidence visible without turning the appendix into a benchmark-ID dump. The pattern is consistent with the paired table: \cheapmain has the strongest fragment distribution, scaffold has positive central tendency with wider tails, and linker remains task-dependent.

\subsection{Claim matrix}
The safe claim matrix used to write the paper is summarized as follows. Strong claims: trust-region delivery activates quotient residuals; deployed residual guidance improves official seed-0 fragment and scaffold validity at modest overhead; \cheapmain improves official seed-0 linker validity while \pcmain is near-neutral; expensive rollout is not the deployed method; novelty and diversity do not collapse in the matched multi-seed molecular slice. Bounded claims: the quality--runtime trade-off is task-dependent; side-chain decoration is a stress test where default ligand-generation budgets over-control the task. Weakened claims: docking is mixed and should not be stated as universal binding improvement; sham residual is close enough on fragment that the completed controls should not be read as ruling out every possible scalarized-guidance alternative.

\subsection{Anti-reparameterization ablation}
\begin{table}[h]
    \caption{Fragment ablation frontier. \teacher is useful as a diagnostic upper-cost point but is not the deployed method. The comparison separates the core trust-region delivery rule from rollout-heavy or unstructured guidance controls.}
    \label{tab:ablation_frontier}
    \centering
    \small
    \begin{tabular}{lcccc}
\toprule
Fragment variant & Seeds & Validity $\uparrow$ & Runtime & Control/base \\
\midrule
\base & 3 & $0.842$ & $1.00\times$ & 0.000 \\
\sectiononly & 1 & $0.896$ & $1.07\times$ & -- \\
\sham & 3 & $0.888$ & $1.22\times$ & 0.255 \\
\pcmain & 3 & $0.896$ & $1.10\times$ & 0.193 \\
\cheapmain & 3 & $\mathbf{0.905}$ & $1.22\times$ & 0.271 \\
\teacher & 3 & $0.898$ & $2.15\times$ & 0.253 \\
\bottomrule
\end{tabular}

\end{table}

\subsection{Side-chain budget stress test}
\begin{figure}[h]
    \centering
    \includegraphics[width=0.96\linewidth]{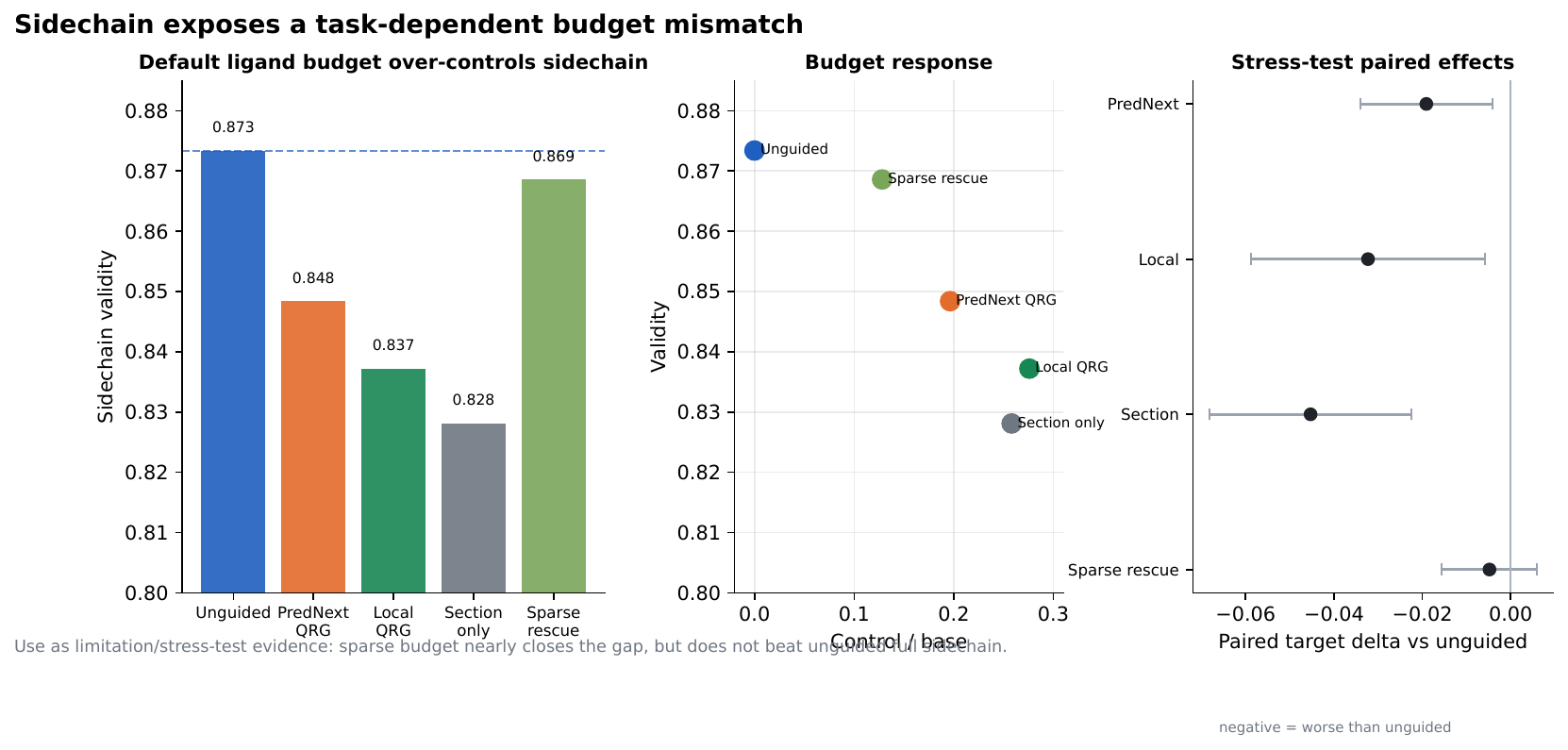}
    \caption{Side-chain decoration exposes a task-dependent budget mismatch. Default ligand-generation budgets reduce side-chain validity relative to the unguided sampler. A sparse low-budget rescue nearly closes the gap, but does not beat unguided in the full sweep.}
    \label{fig:sidechain_stress}
\end{figure}

\subsection{Corrected docking diagnostic}
\begin{table}[h]
    \caption{Corrected raw-pocket docking diagnostic. Values are paired mean differences versus the base sampler with bootstrap intervals. Negative means the guided variant has better raw \Vina score. The signal is clearest on scaffold hopping and mixed on the other tasks.}
    \label{tab:vina_summary}
    \centering
    \small
    \begin{tabular}{lccc}
\toprule
Task / method & Paired targets & Mean $\Delta$Vina $\downarrow$ & Better targets \\
\midrule
Fragment / \pcmain & 30 & $0.061\;[-0.049,0.174]$ & 14/30 \\
Fragment / \cheapmain & 30 & $-0.101\;[-0.189,-0.023]$ & 17/30 \\
Scaffold / \pcmain & 44 & $-0.132\;[-0.234,-0.035]$ & 28/44 \\
Scaffold / \cheapmain & 44 & $\mathbf{-0.285}\;[-0.397,-0.178]$ & 35/44 \\
Linker / \pcmain & 42 & $-0.001\;[-0.087,0.081]$ & 20/42 \\
Linker / \cheapmain & 42 & $0.038\;[-0.055,0.130]$ & 17/42 \\
\bottomrule
\end{tabular}

\end{table}

\section{Broader impact and compute resources}\label{app:impact_compute}

\subsection{Broader impact}
\qrg is a sampling-time method for scientific molecular generation. Its intended use is methodological: evaluating how quotient-aware guidance can improve frozen generative samplers under explicit runtime budgets. The generated molecules in this paper are computational candidates produced inside benchmark pockets; they are not validated binders, drugs, or therapeutic leads. We therefore separate generation validity from docking diagnostics and avoid claims of biological activity. This boundary is important because over-interpreting raw docking scores or visually plausible structures could mislead downstream drug-design decisions.

The positive impact of this work is improved evaluation discipline for inference-time molecular guidance: the method reports wall-clock cost, delivered control budget, validity, novelty, diversity, and task-dependent failures rather than only best-case samples. Potential negative impact comes from the general dual-use nature of molecular generation. The work does not release a new biological assay, trained wet-lab model, or actionable synthesis pipeline; nonetheless, any use of generated molecules should require independent chemistry review, safety screening, and experimental validation.

\subsection{Compute resources}
All real molecular experiments use a frozen \TargetDiff backbone inside the \CBGBench evaluation code. Official seed-0 sweeps are split into 20-target shards, with 100 generated samples per target, batch size 2, and seed 0. Fragment, scaffold, linker, and side-chain official tasks use the same sharding contract. Runtime is reported as wall-clock seconds per generated sample from each run's \texttt{run\_summary.json}; aggregate rows weight by the number of generated samples. Contaminated local co-runs are excluded from fair runtime comparisons when clean single-run measurements are available.

The official ligand-generation/editing shards run on cluster GPU workers. Completed official shard durations for linker are typically about 3.6--5.2 hours per shard, with occasional slower shards around 7 hours. Fragment and scaffold official rows report aggregate seconds/sample in \cref{tab:molecular_main}; these values include denoising, guidance, postprocessing, and file writing as captured by the benchmark runner. The multi-seed molecular slice and sham/teacher controls required additional runs, with rollout-teacher variants intentionally treated as diagnostic high-cost points rather than deployed methods.

\begin{table}[h]
    \caption{Completed evaluation inventory. The table records what was completed and how each asset is used in the paper. It is an accounting table, not an additional performance ranking.}
    \label{tab:completed_inventory}
    \centering
    \small
    \resizebox{\linewidth}{!}{\begin{tabular}{lccp{0.50\linewidth}}
\toprule
Evaluation asset & Completed rows & Generated samples & Role in the paper \\
\midrule
Official ligand-generation/editing sweeps & 9 & 50{,}400 & Main validity, runtime, and control-budget evidence for fragment growing, scaffold hopping, and linker design in \cref{tab:molecular_main}. \\
Official section-only controls & 3 & 16{,}800 & Residual-off controls used to separate section delivery from quotient-residual guidance. \\
Official side-chain sweep & 4 & 25{,}600 & Out-of-family budget-transfer stress test; not counted as a ligand-generation win claim. \\
Matched multi-seed molecular slice & 40 & -- & Novelty/diversity preservation, development ablations, rollout-teacher diagnostics, and sham-control comparisons. \\
Target-balanced docking diagnostic & 6 & -- & Supporting raw-pocket \Vina diagnostic with paired target comparisons; not a binding-affinity claim. \\
\midrule
Official full aggregate total & 16 & 92{,}800 & Completed official seed-0 aggregate rows across all four molecular tasks and four reported method families. \\
\bottomrule
\end{tabular}
}
\end{table}

\begin{table}[h]
    \caption{Compute accounting for the official seed-0 ligand-generation/editing rows that carry the main molecular claim. Approximate GPU hours are computed from generated samples times aggregate wall-clock seconds/sample divided by 3600, and therefore include denoising, guidance, postprocessing, and file writing as recorded by the runner.}
    \label{tab:compute_accounting}
    \centering
    \small
    \begin{tabular}{llrrrr}
\toprule
Task & Method & Generated & sec/sample & Approx. GPU h & Control/base \\
\midrule
Fragment & \base & 6{,}100 & 16.68 & 28.3 & 0.000 \\
Fragment & \pcmain & 6{,}100 & 18.28 & 31.0 & 0.192 \\
Fragment & \cheapmain & 6{,}100 & 19.91 & 33.7 & 0.265 \\
Scaffold & \base & 6{,}400 & 17.99 & 32.0 & 0.000 \\
Scaffold & \pcmain & 6{,}400 & 20.36 & 36.2 & 0.194 \\
Scaffold & \cheapmain & 6{,}400 & 20.18 & 35.9 & 0.268 \\
Linker & \base & 4{,}300 & 18.82 & 22.5 & 0.000 \\
Linker & \pcmain & 4{,}300 & 19.72 & 23.6 & 0.183 \\
Linker & \cheapmain & 4{,}300 & 21.38 & 25.5 & 0.251 \\
\bottomrule
\end{tabular}

\end{table}

\subsection{Denoiser-query accounting}
The deployed methods are budgeted inference-time corrections rather than many-rollout optimizers. \Cref{tab:query_accounting} summarizes the guidance schedule used in the reported configuration. The counts are guidance-delivery opportunities under the 500-step sampler schedule; they are not meant as hardware-level FLOP measurements. The key distinction is that \pcmain and \cheapmain avoid the rollout-teacher query pattern used only for diagnostics.

\begin{table}[h]
    \caption{Denoiser-query and guidance-call accounting. The table separates deployed QRG variants from rollout-teacher diagnostics and explains why the reported runtime overhead is modest relative to a rollout-heavy inference scheme.}
    \label{tab:query_accounting}
    \centering
    \small
    \resizebox{\linewidth}{!}{\begin{tabular}{lp{0.25\linewidth}cp{0.38\linewidth}}
\toprule
Method family & Guidance schedule & Guidance deliveries/sample & Denoiser-query interpretation \\
\midrule
\base & Frozen reverse sampler only & 0 & No guidance branch and no extra denoiser rollout. \\
\sectiononly & Section branch, residual budget zero & 500 & Tests delivery without quotient-residual correction. \\
\pcmain & Predicted-next residual every two denoising steps & 250 & Reuses the sampler's predicted next state; deployed point avoids rollout-teacher queries. \\
\cheapmain & Local-radius quotient surrogate with trust-region delivery & 500 & Uses a cheap local surrogate rather than querying a rollout teacher. \\
\teacher & One shared rollout before residual evaluation & 500 & Diagnostic high-cost teacher/oracle point, not the deployed method. \\
\bottomrule
\end{tabular}
}
\end{table}

\subsection{Existing assets and licenses}
The experiments reuse the public \TargetDiff model family and the \CBGBench benchmark/task definitions, both cited in the main text. The upstream \TargetDiff repository identifies its code license as MIT. The upstream \CBGBench repository identifies its code license as GPL-3.0, and the \CBGBench OpenReview paper page identifies the paper license as CC BY 4.0. Our code supplement contains the QRG implementation, runner wrapper, aggregation scripts, and configuration description; it does not redistribute upstream benchmark data, checkpoints, generated molecules, logs, raw structures, or rendered figures, and it does not assert a new license over upstream benchmark assets. Users should obtain any upstream data and checkpoints from the original sources under those original terms.

\section{Reproducibility notes}\label{app:repro}

\paragraph{Code supplement.}
The supplementary code package is intentionally minimal. It contains the QRG implementation, the frozen \TargetDiff/\CBGBench runner wrapper, runtime and metric summarization scripts, and an environment file. It does not redistribute raw benchmark data, processed data, generated molecules, checkpoints, logs, rendered figures, manuscript source, or precomputed result tables. Full regeneration therefore requires obtaining the upstream \CBGBench/\TargetDiff assets under their original terms.

\paragraph{Reproducing the molecular evaluation.}
The benchmark runner uses the frozen \TargetDiff checkpoint, \CBGBench task configuration, the official task split, and the guidance parameters listed in \cref{app:expdetails}. Official sweeps are generated as 20-target shards with 100 samples per target and batch size 2. The aggregate scripts read each shard's \texttt{run\_summary.json} and generated SDF outputs, then recompute validity, runtime, control/base ratio, chemistry metrics, side-chain summaries, and corrected docking diagnostics.

\end{document}